\documentclass[journal]{IEEEtran}

\usepackage{cite}
\usepackage{amsmath,amssymb,amsfonts}
\usepackage{algorithmic}
\usepackage{graphicx}
\usepackage{textcomp}
\usepackage{bm}
\usepackage{makecell}
\usepackage[normalem]{ulem}
\usepackage{xcolor}
\usepackage{tabularx,threeparttable}
\usepackage[ruled,linesnumbered]{algorithm2e}
\usepackage{booktabs}
\usepackage{arydshln}
\usepackage{comment}
\newcommand{\yj}[1]{\textcolor{black}{#1}}
\newcommand{\sj}[1]{\textcolor{black}{#1}}
\begin{document}
\title{Free-DyGS: Camera-Pose-Free Scene Reconstruction for Dynamic Surgical Videos \\ with Gaussian Splatting}
\author{Qian Li, Shuojue Yang, Daiyun Shen, Jimmy Bok Yan So, \\Jing Qin, \IEEEmembership{Member, IEEE} and Yueming Jin, \IEEEmembership{Member, IEEE}
%\thanks{{This paragraph of the first footnote will contain the date on which
% you submitted your paper for review. It will also contain support information,
% including sponsor and financial support acknowledgment. For example, 
% ``This work was supported in part by the U.S. Department of Commerce under Grant BS123456.'' }
%\thanks{This work was supported by Ministry of Education Tier 1 grant, National University of Singapore (NUS), Singapore (A-8001946-00-00).}
\thanks{Q. Li, S. Yang, D. Shen, and Y. Jin are with Department of Biomedical Engineering, National University of Singapore (NUS), Singapore. Y. Jin is also with Department of Electrical and Computer Engineering, NUS.}
\thanks{Jimmy B.Y. So is with Yong Loo Lin School of Medicine, NUS, and Department of Surgery, National University Hospital. J. Qin is with the Centre for Smart Health, The Hong Kong Polytechnic University.}
\thanks{Corresponding author: Yueming Jin. (e-mail: ymjin@nus.edu.sg)}
}

\maketitle

\begin{abstract}
High-fidelity reconstruction of surgical scene is a fundamentally crucial task to support many applications, such as intra-operative navigation and surgical education. 
However, most existing methods assume the ideal surgical scenarios - either focus on dynamic reconstruction with deforming tissue yet assuming a given fixed camera pose, or allow endoscope movement yet reconstructing the static scenes.
In this paper, we target at a more realistic yet challenging setup - free-pose reconstruction with a moving camera for highly dynamic surgical scenes. 
Meanwhile, we take the first step to introduce Gaussian Splitting (GS) technique to tackle this challenging setting and propose a novel GS-based framework for fast reconstruction, termed \textit{Free-DyGS}. 
Concretely, our model embraces a novel scene initialization in which a pre-trained Sparse Gaussian Regressor (SGR) can efficiently parameterize the initial attributes. For each subsequent frame, we propose to jointly optimize the deformation model and 6D camera poses in a frame-by-frame manner, easing training given the limited deformation differences between consecutive frames. A Scene Expansion scheme is followed to expand the GS model for the unseen regions introduced by the moving camera. Moreover, the framework is equipped with a novel Retrospective Deformation Recapitulation (RDR) strategy to preserve the entire-clip deformations throughout the frame-by-frame training scheme.
The efficacy of the proposed Free-DyGS is substantiated through extensive experiments on two datasets: StereoMIS and Hamlyn datasets. The experimental outcomes underscore that Free-DyGS surpasses other advanced methods in both rendering accuracy and efficiency. Code will be available.
\end{abstract}


\section{Introduction}
\label{sec:introduction}
\begin{figure*}[t]
	\centering
	\includegraphics[width=0.95\textwidth]{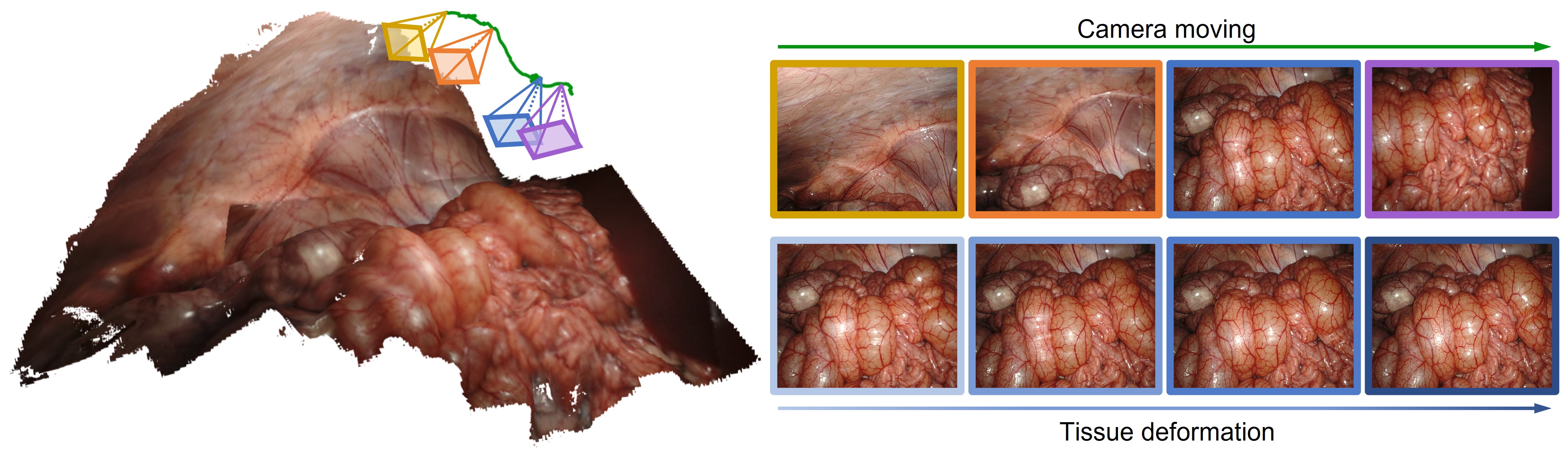}
	\vspace{-3mm}
	\caption{Point clouds (left) reconstructed by our approach from StereoMIS dataset with camera trajectory estimation (green curve) and rendered images. Images within the colored frames (top right) illustrate the scene captured under various camera poses. Those within the gradient blue frames (bottom right) displays the tissue deformation. } 
	\label{fig:overview}
	\vspace{-5mm}
\end{figure*}
\IEEEPARstart{R}{econstructing} surgical scenes from laparoscopic and endoscopic videos has profound implications for enhancing visualization quality and improving the safety of surgical procedures. It can improve the surgical experience by augmenting the surgeon's perception of the operating field~\cite{bianchi2020impact}, and facilitating the identification of critical structures such as blood vessels and tumors~\cite{schiavina2019three,jin2022exploring}. This advancement also facilitates multi-view observation of the surgical scene during procedures~\cite{otsuki2023surgical}, which can support higher-level downstream tasks including collaborative surgery, safety monitoring, and skill assessment. Moreover, the reconstructed virtual simulation environment can offer an education platform for surgical trainees~\cite{lange2000virtual}. Surgical scene reconstruction will also pave the way for integrating Augmented Reality technologies~\cite{tang2018augmented,chen2018slam}, enriching the interactive experience of medical professionals during surgery or training sessions. Furthermore, 3D reconstruction is a crucial building block for registration between pre-operative images and intra-operative videos, further providing precise navigation to enhance surgical procedures~\cite{zha2023endosurf}.

However, surgical video reconstruction presents significant challenges, given the simultaneous occurrence of camera movement and tissue deformation during surgery (cf. Fig. \ref{fig:overview}). Specifically, i) in contrast to natural scenes, tissue deformation needs to be considered in surgical scene reconstruction, which is an inherent characteristic that encompasses phenomena such as respiration, cardiac pulsation, intestinal motility, and interactions with surgical instruments. ii) Camera motion is common during surgical procedures, as surgeons frequently adjust the view to observe different regions. However, accurate camera pose, which is essential for scene reconstruction, is typically not directly accessible in surgeries.

Though various studies have been proposed for surgical scene reconstruction, most of them either focus on deformation modeling while assuming a fixed camera pose, or investigate reconstruction under camera movement while ignoring scene dynamics (i.e., tissue deformations). For example, some recent approaches~\cite{huang2024endo, liu2024endogaussian} focus on advanced deformation modeling by using Gaussian splatting (GS), however, they assume that the camera pose is fixed and needs to be provided. Others are proposed for unposed scene reconstruction while assuming a static scene~\cite{wang2024endogslam, guo2024free, yugay2023gaussian}. They propose to estimate camera motion by comparing the reconstructed scene with incoming frames. However, when tackling dynamic surgical scenarios, the reconstructed static scenes from these methods fail to match the deformed tissues, which can lead to incorrect pose estimation and blurred reconstruction. 
Overall, most existing methods fail to simultaneously address the two critical challenges of surgical scene reconstruction and, therefore, struggle to adapt to real surgical practice with suboptimal reconstruction performance.

To the best of our knowledge, the only study that considers both tissue deformation and camera movement for surgical scene reconstruction is Flex~\cite{stilz2024flex}, a recently proposed method based on NeRF. It utilizes a progressive optimization scheme to jointly reconstruct the scene and estimate camera poses from scratch. This approach however takes excessively long time (a few hours) to reconstruct a video clip (xx seconds). The underlying reasons are that Flex relies on MLP-based NeRF requiring a long training time, meanwhile, it reconstructs frames by exploiting a randomly selecting training strategy from the entire sequence, which may lead to large discrepancies between the current and previously reconstructed frames, further retarding model convergence. Additionally, the lack of efficient initialization methods also limits its capability for rapid reconstruction.

\sj{In this paper, we take the first step to investigate the potential of GS technique for fast surgical scene reconstruction under this more realistic yet challenging setting, considering \textbf{dy}namic tissue deformation under \textbf{free} camera pose condition. 
We present a novel GS-based framework, named \textbf{Free-DyGS}, which embraces three core components, including an efficient initializer, a new joint learning scheme with scene expansion, and a retrospective deformation recapitulation strategy, to enable GS to yield its efficacy for unposed surgical scene reconstruction with high efficiency.
Specifically, to accelerate the GS reconstruction, our framework starts with a novel scene initialization where a pre-trained Sparse Gaussian Regressor (SGR) efficiently parameterizes the initial Gaussian attributes from the first RGBD frame in a single feedforward pass. 
For subsequent frames with tissue deformations and unknown camera extrinsics, each incoming frame is first partially represented by deforming the Gaussian points reconstructed from previous frames. We therefore propose jointly optimizing the deformation model and 6D camera poses in a frame-by-frame sequential manner. In this regard, the incoming frame encompasses limited deformation/scene difference from the previous frame, therefore facilitating the point propagation and training process. 
Meanwhile, as the camera moves, the new frame inevitably contains unseen regions, we design a Scene Expansion scheme that re-utilizes SGR to efficiently extend the GS model to unseen regions, which also enhances reconstruction quality. 
Moreover, since the deformation model is shared across frames and may exhibit long-range forgetting, we propose a Retrospective Deformation Recapitulation (RDR) strategy with a temporal decoupling deformation model to recall earlier frames’ deformations during training, effectively preserving historical information.
}

Our main contributions can be summarized as follows:
\begin{enumerate}
\item
To our best knowledge, Free-DyGS is the first GS-based framework for fast reconstruction with unknown camera motions and complex tissue deformations. Free-DyGS proposes to jointly optimize camera positioning and deformation modeling by a novel frame-by-frame scheme with scene expansion.

\item
We introduce a sparse Gaussian regressor module to efficiently parameterize Gaussian attributes in a single feedforward for scene initialization and expansion, which can reduce training time for high-quality reconstruction. 

\item
We develop a retrospective deformation recapitulation strategy, which includes temporal decoupling deformation model and retrospective learning, to preserve deformation model to cover the entire clip during the frame-by-frame training.

\item Extensive experiments on the StereoMIS and Hamlyn datasets demonstrate the effectiveness of our method in reconstructing deformable scenes, outperforming other state-of-the-art techniques in terms of both reconstruction quality and training efficiency.

\end{enumerate}

\section{Related Works}

% \noindent \textbf{Traditional scene reconstruction methods}
Reconstructing a 3D surgical scene from a collection of 2D images is a prevalent and significant task across various applications. Traditional approaches aim to represent the scene as a 3D point cloud. SfM techniques, such as COLMAP~\cite{schoenberger2016sfm}, are employed to deduce the camera pose, which is then utilized to integrate the point clouds associated with each frame. SLAM-based methods~\cite{taketomi2017visual} are able to track the camera and map the environment simultaneously. These methodologies have been applied in endoscopic reconstruction tasks, as demonstrated in \cite{zhou2019real, song2017dynamic, zhou2021emdq}. However, these methods usually assume a static scene and may fail when applied to deformable scenes.

% \noindent \textbf{Dynamic scene reconstruction}
In recent years, neural radiance fields (NeRF)~\cite{mildenhall2021nerf} have emerged as a prominent approach for reconstructing static scenes. Building upon it, EndoNeRF~\cite{wang2022neural }, LerPlane~\cite{yang2023neural}, and ForPlane~\cite{yang2023efficient} introduced a time-variant neural displacement field to represent dynamic surgical scenes. 
Furthermore, the advanced 3D GS techniques are emerging~\cite{kerbl20233d}. Several GS-based methods, such as EndoGaussian~\cite{liu2024endogaussian}, Endo-GS~\cite{huang2024endo}, and Deform3DGS~\cite{yang2024deform3dgs} are proposed for dynamic surgical scene reconstruction with fast speed. However, most of these methodologies can only work on scenes with fixed cameras.

% \noindent \textbf{Camera pose estimation}
To enhance reconstruction techniques, several approaches have been introduced to explore scene reconstruction without relying on accurate camera poses in nature domain. Early works based on NeRF~\cite{wang2021nerf,lin2021barf,bian2023nope} refine camera poses by leveraging the color discrepancy between rendered and original images. As an extension, COLMAP-free GS~\cite{Fu_2024_CVPR} adopts GS framework to achieve scene reconstruction without requiring camera poses. More advanced methods~\cite{yugay2023gaussian,yan2024gs,li2024sgs} propose SLAM strategies incorporating GS. Recently, EndoGSLAM~\cite{wang2024endogslam} and Free-SurGS~\cite{guo2024free} estimate camera poses and extend GS to surgical scene reconstruction. However, these methods are primarily designed for static scenes and may not be well-suited for dynamic surgical reconstruction tasks.

% \noindent \textbf{Pose-free dynamic scene reconstruction}
Moreover, methods to address a more challenging task on dynamic scene reconstruction without camera poses have been emerging. RoDyNeRF~\cite{liu2023robust} aims to simultaneously optimize camera poses and reconstruct natural dynamic scenes. It models the scene as a composite of static and dynamic parts for camera motion optimization and moving object modeling, respectively. However, it may not be well-suited for surgical scene reconstruction due to the high deformability and low rigidity inherent in such environments~\cite{saha2023based}.
Flex~\cite{stilz2024flex}, as an advancement in the medical domain, also leverages NeRF as its base technology and jointly learns the camera pose tissue deformation during training. Nonetheless, similar to other NeRF-based techniques, it typically requires an extensive training time of several hours, which can significantly impede their practical clinical utility.

\section{Methods}
\begin{figure*}[t]
	\centering
	\includegraphics[width=0.9\textwidth]{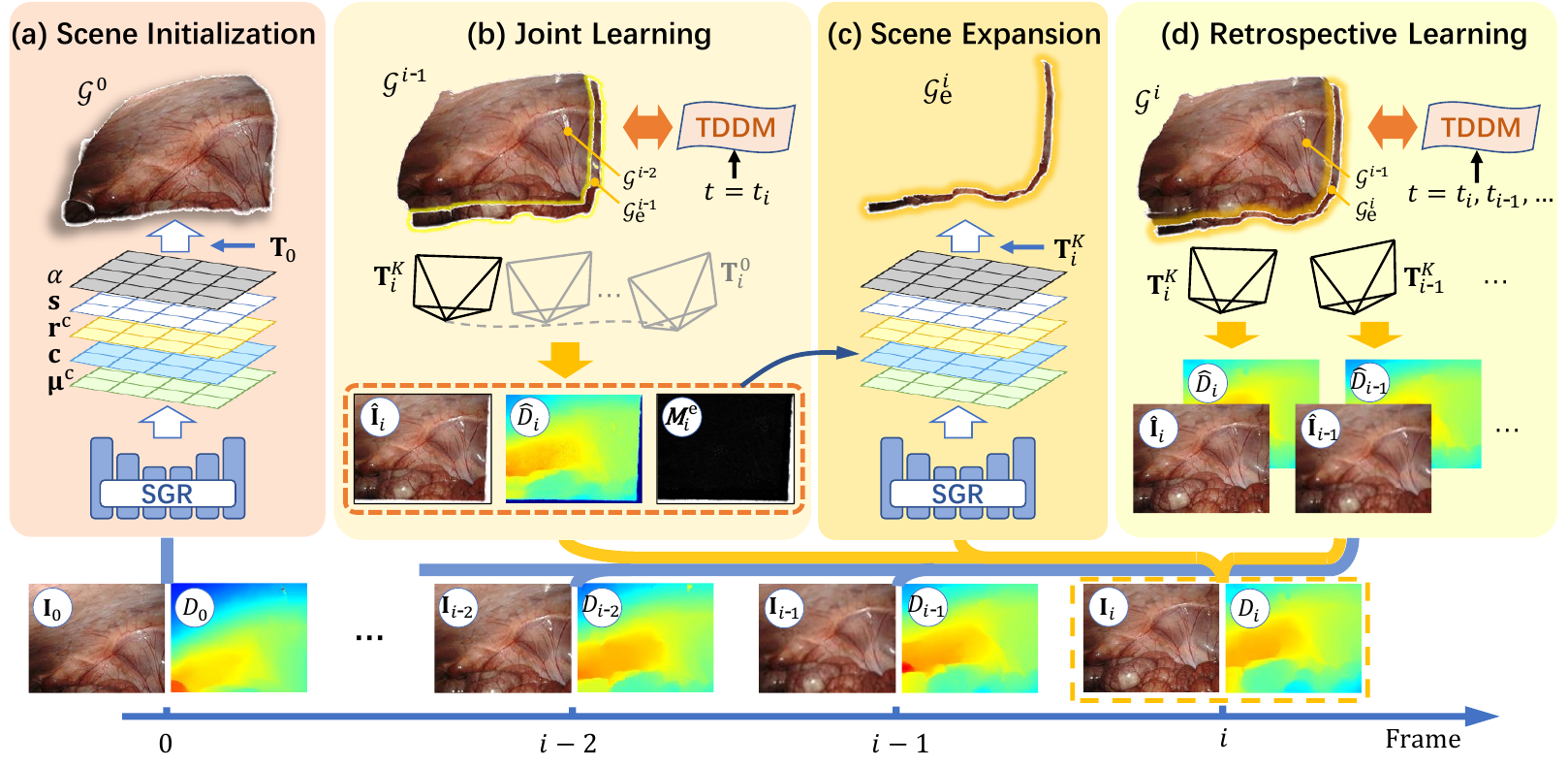}
	\vspace{-3mm}
	\caption{Overview of the proposed Free-DyGS framework. The framework starts with (a) Scene initialization with SGR, predicting a well-parameterized Gaussian canonical model. For subsequent frames, Free-DyGS proposes to (b) jointly optimize camera positioning and deformation modeling by the novel frame-by-frame training scheme, with (c) Scene Expansion to introduce new Gaussian points for unseen regions. (d) Retrospective Deformation Recapitulation strategy is finally designed to mitigate long-range forgetting in deformation modeling.}

	\label{fig:method}
	\vspace{-3mm}
\end{figure*}

The overview of our Free-DyGS is illustrated in Fig. \ref{fig:method} and Algorithm \ref{alg:frame}. 
In this section, we first briefly introduce the GS technique preliminaries (Sec. \ref{sec:3.1}) and then describe the details of the proposed Free-DyGS. Our approach begins with an efficient scene initialization to predict a well-parameterized Gaussian canonical model from the initial frame (Sec. \ref{sec:3.2}). 
For subsequent frames, we propose a frame-by-frame training paradigm to jointly estimate the tissue deformation and the camera pose, followed by a scene expansion scheme to introduce new Gaussian points to represent the unseen regions (Sec. \ref{sec:3.3}).
Lastly, we develop a retrospective deformation recapitulation strategy with a temporal decoupling deformation model, to alleviate the long-range forgetting issue of deformation modeling (Sec. \ref{sec:3.4}).

\newcommand\mycommfont[1]{\footnotesize\ttfamily\textcolor{blue}{#1}}
\SetCommentSty{mycommfont}
\begin{algorithm}[t]
\caption{The proposed Free-DyGS framework.}\label{alg:frame}
\KwIn{Frame sequence datasets $\{(t_i, \mathbf{I}_i, \mathbf{D}_i)\}_{i=0}^{I-1}$, Pre-trained SGR network $SGR$, Retrospective window width $\omega$}

\KwOut{Scene reconstruction $\mathcal{G}^I$, Trained TDDM $\Phi$.}
Initialize TDDM $\Phi$ with $t_{I-1}$\;
\tcc{Scene Initialization via SGR}
Initialize $\mathbf{T}_0=\mathbf{E}$\;
$ \mathcal{G}^0:=tf({SGR}(\mathbf{I}_0,\mathbf{D}_0),\mathbf{T}_0)$\;
\tcc{frame-by-frame training}
\For{$i = 1,2,...,I-1$}{
    \tcc{Joint learning with scene expansion}
    Initialize $\mathbf{T}_i^{0}=\mathbf{T}_{i-1}^{K}V(\{\mathbf{T}_{i-l}^{K}\}_{l=0}^{2L-1})$\;
    \For{$k = 1,2,...,K$}{
        $\mathbf{T}^k_i, \Phi \leftarrow train(\mathbf{I}_i, \mathbf{D}_i, \mathcal{G}^{i-1}+\Phi(t_i), \mathbf{T}_i^{k-1})$\;
    }
    %\tcc{Scene expansion}
    $M^{\text{e}}_i=opacity\_render(\mathcal{G}^{i-1}+\Phi(t_i), \mathbf{T}_i^{K})$\;
    $\mathcal{G}_{\text{e}}^i:=tf(M^{\text{e}}_i \odot {SGR}(\mathbf{I}_i,\mathbf{D}_i),\mathbf{T}^K_i)$\;
    $\mathcal{G}^i=\mathcal{G}^{i-1}\oplus \mathcal{G}_{\text{e}}^i$\;
    \tcc{RDR learning}
    Randomly select training indices $\Omega$ from $ (i-\omega, i]$\;
    \For{$j$ in $\Omega$}{
        $\Phi \leftarrow train(\mathbf{I}_j, \mathbf{D}_j, \mathcal{G}^{i}+\Phi(t_j), \mathbf{T}_j^{K})$\;
    }
        
}
        
\end{algorithm}

\subsection{Preliminaries for 3D Gaussian Splatting}
\label{sec:3.1}
Gaussian splatting~\cite{kerbl20233d} is an emerging technique for fast 3D reconstruction of static scenes. It utilizes a collection of generalized Gaussian point clouds to represent the scene. Each Gaussian point, denoted as \( G_n \), encompasses several attributes: position \( \bm{\mu}_n \), scale \( \mathbf{s}_n \), rotation quaternion \( \mathbf{r}_n \), opacity $\alpha_n$ and color $\mathbf{c}_n$ described by spherical harmonics (SH) parameters. The position and the geometric shape can be mathematically expressed with the 3D coordinates $\mathbf{x}$ as
\begin{equation}
G_n(\mathbf{x})=\exp{\left(-\frac{1}{2}(\mathbf{x}-\bm{\mu}_n)^T\mathbf{\Sigma}_n^{-1}(\mathbf{x}-\bm{\mu}_n)\right)}
\end{equation}
where $\mathbf{\Sigma}_n=\mathbf{R}_n\mathbf{S}_n\mathbf{S}_n^T\mathbf{R}_n^T$ is the covariance matrix, $\mathbf{R}_n$ is the rotation matrix derived from the quaternions $\mathbf{r}_n$
, and $\mathbf{S}_n$ the diagonal matrix of the scaling vector $\mathbf{s}_n$.

Given a camera pose matrix $\mathbf{T}$, these points can then be projected on the image plane.  $\alpha$-blending~\cite{kerbl20233d} can further be performed to render colored images $\hat{\mathbf{I}}$ and depth map $\hat{\mathbf{D}}$.

\subsection{Efficient Scene Initialization}
\label{sec:3.2}
At the beginning of our framework, obtaining accurate scene initialization is crucial, as it provides essential geometric and texture priors to facilitate the reconstruction of new frames or scenes. We propose a Gaussian parameterization method to efficiently initialize the scene.

\textbf{Sparse Gaussian Regressor.}
Our framework takes an RGB image $\mathbf{I}\in\mathbb{R}^{(H \times W)\times3}$ and the corresponding depth map $\mathbf{D} \in \mathbb{R}^{H \times W}$ as input and is trained to predict pixel-aligned attribute maps where each pixel represents a Gaussian primitive with pixel-level attributes. 
However, pixel-aligned Gaussian attribute maps with high resolution lead to excessively dense Gaussian points, impairing reconstruction efficiency and increasing computational costs. 

\begin{figure}[t]
	\centering
	\includegraphics[width=0.43\textwidth]{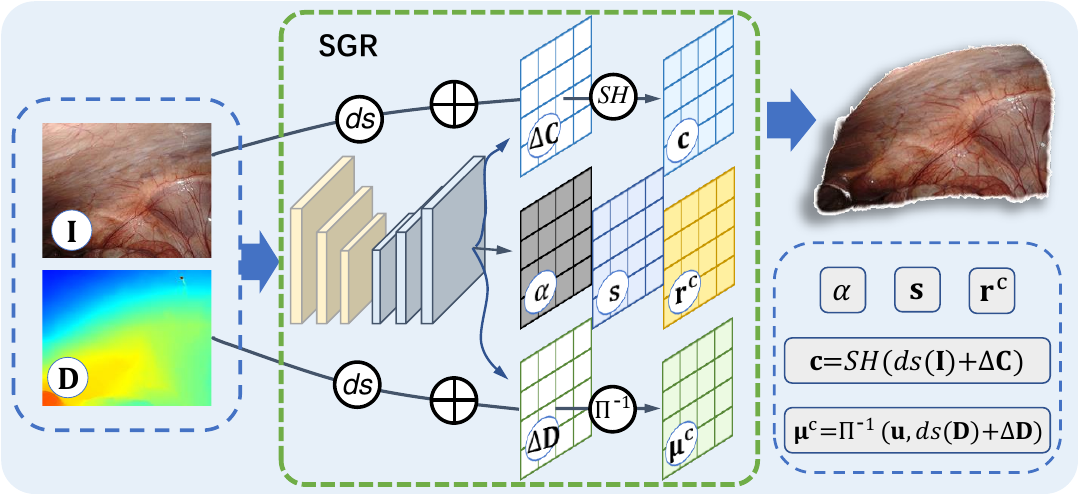}
	% \vspace{-3mm}
	\caption{Illustration of proposed Sparse Gaussian Regressor module, which is designed to produce pixel-aligned Gaussian attribute maps ($\alpha$, $\mathbf{s}$, $\mathbf{r^{\text{c}}}$) and calibration terms ($\Delta C$, $\Delta \mathbf{D}$) from the input frame  $\{\mathbf{I}, \mathbf{D}\}$.} 
	\label{fig:generalizable}
\end{figure}

We therefore propose a CNN-based module, Sparse Gaussian Regressor (SGR), to generate well-parameterized Gaussian points for initialization. 
As shown in Fig. \ref{fig:generalizable}, we first downsample RGB-D images at scale $s$ before feeding them into the Gaussian regressor, which decreases the pixel numbers and produces sparse Gaussian attribute maps with a size of $H/s \times W/s$. Considering that sparse Gaussian point cloud may not accurately represent the scene using original-resolution depth and color maps, we propose to develop additional heads for updating color and depth by predicting two calibration terms \( \Delta\mathbf{C}, \Delta \mathbf{D} \) to refine the sparse attribute maps, 
%其他方法本来是用2个head分别predict c&\mu吗？这里的low-resolution attribute maps是怎么来的呢，在图3中也看不出来
along with three other heads for Gaussian attributes $\alpha$, $\mathbf{s}$, and $\mathbf{r}^\text{c}$ in camera coordinate system $c$. Finally, the pixel-aligned Gaussian color $\mathbf{c}$ and position $\bm{\mu}^\text{c}$ are derived from the calibrated pixel color and the calibrated depth, respectively.
\vspace{-2mm}
\begin{align}
    \mathbf{c}&=SH(ds(\mathbf{I})+\Delta \mathbf{C})\\
    \bm{\mu}^\text{c}&=\Pi^{-1}(\mathbf{u},ds( \mathbf{D})+\Delta \mathbf{D})
\end{align}
where $SH$ is the spherical harmonics converting function, $\Pi^{-1}$ represents the inverse process of projection, $ds$ is the downsampling operation, and $\mathbf{u} \in\mathbb{R}^{(H/s\times W/s)\times 3}$ denotes the downsampled pixel coordinates. \yj{Note that we pre-train the SGR offline by hold-out surgical data, which can achieve impressive rendering quality without online refinement.}

\textbf{Scene Initialization.}
By default, the first frame initiates the Gaussian point cloud. Given the image $\mathbf{I}_0$ and depth map  $\mathbf{D}_0$, SGR predicts Gaussian attributes,
\begin{equation}
\{\alpha, \mathbf{c}, \mathbf{s}, \mathbf{r}^\text{c}, \bm{\mu}^\text{c}\}=SGR(\mathbf{I}_0,\mathbf{D}_0)
\end{equation}
where $\mathbf{r}^\text{c}$ and $\bm{\mu}^\text{c}$ are established on the camera coordinate system. They are then transformed into the world coordinate system using the initial camera pose \( \mathbf{T}_0 \) (if it is not available, $\mathbf{T}_0:= \mathbf{E}$ and the world coordinate system will be set on the first camera coordinate system), generating the initial Gaussians $\mathcal{G}^0$,

\begin{equation}
\begin{aligned}
    \mathcal{G}^0:=&tf(SGR(\mathbf{I}_0,\mathbf{D}_0),\mathbf{T}_0)\\
    =&\{\alpha, \mathbf{c}, \mathbf{s}, Q(\mathbf{T}_0)\otimes\mathbf{r}^\text{c}, \mathbf{T}_0\bm{\mu}^\text{c} \}
    \end{aligned}
\end{equation}
where $tf(\cdot)$ is the attribute maps transformation function, $Q(\cdot)$ represents the function that converts a transformation matrix into a quaternion, and $\otimes$ represents quaternion multiplication.

\subsection{Joint Camera Positioning and Reconstruction}
\label{sec:3.3}

After initializing the canonical Gaussian model $\mathcal{G}^{0}$, our framework sequentially processes subsequent video frames $\{(\mathbf{I}_1, \mathbf{D}_1),\;...,(\mathbf{I}_i, \mathbf{D}_i),\;...\}$ to model scene deformations and localize endoscopic camera. Considering that tissue deformations and camera motions simultaneously occur during the surgery, we thus propose a novel optimization strategy to learn per-frame camera pose and scene deformation jointly. Specifically, for the $i$-th frame, our framework first initializes a pose \(\mathbf{T}^0_i \) by aggregating poses estimated in previous $2L$ frames $\{\mathbf{T}_{i-l}\}_{l=1}^{2L}$ into an approximated velocity \( \mathbf{V}_i \). This velocity is calculated in the Lie-algebra domain like~\cite{wang2023bad}:
\begin{equation}
    \mathbf{T}^0_i=\mathbf{T}_{i-1}\mathbf{V}_i
\end{equation}
\vspace{-5mm}
\begin{equation}
    \mathbf{V}_i = \exp(\sum_{l=1}^{L}\log(\mathbf{T}_{i-l}\mathbf{T}^{-1}_{i-L-l})/L^2) 
\end{equation}
where $\mathbf{T}_i\in\mathbf{SE}(3)$, $\text{exp}(\cdot)$ and $\text{log}(\cdot)$ denote the exponential and logarithm mapping for Lie algebra and group, respectively.

Given this initialized camera pose \(\mathbf{T}^0_i \) for the $i$-th frame, our framework leverages the previously reconstructed canonical Gaussian model $\mathcal{G}^{i-1}$ to iteratively optimize this pose and current-frame deformations. In our work, the deformed Gaussian $\tilde{\mathcal{G}}^{i-1}$ is defined by adding the variation in attributes to the canonical Gaussian attributes, formulated as:
\begin{equation}
    \tilde{\mathcal{G}}^{i-1}_{t_{i}} = \mathcal{G}^{i-1} + \Phi(t_i)
    \label{deform_gs}
\end{equation}   
where $t_i$ is the timestamp of the current $i$-th frame; $\Phi(t_i)$ represents the proposed TDDM (details in Sec.~\ref{sec:3.4} 1)) that outputs variations in per-Gaussian attributes at timestamp $t_i$, and `$+$' represents the numerical addition across attributes. This joint learning phase proceeds $K$ iterations, where $K$ is empirically set to 10 in our work. In the $k$-th iteration, the iterative camera pose \(\mathbf{T}^k_i \) and deformation model $\Phi$ is updated toward minimizing the difference between the ground-truth frame $\{\bm{I}_i,\;\bm{D}_i\}$ and corresponding RGB-D maps $\{\hat{\bm{I}}_i,\;\hat{\bm{D}}_i\}$ rendered by deformed Gaussian model $\tilde{\mathcal{G}}^{i-1}_{t_{i}}$ with the current camera pose \(\mathbf{T}^k_i \). 

\textbf{Scene Expansion.} In addition to the joint learning that explicitly models the deformed Gaussian $\tilde{\mathcal{G}}^{i-1}_{t_{i}}$, our framework also progressively incorporates new Gaussian points to canonical Gaussian $\mathcal{G}$. This is because camera motion inevitably introduces unseen regions of the surgical site into the current view and these un-initialized regions can complicate reconstruction, which further degrades the reconstruction quality. To handle this, given the optimized camera pose \(\mathbf{T}_i \) and tissue deformation model $\Phi$, an opacity map is rendered by deformed Gaussian model $\tilde{\mathcal{G}}^{i-1}_{t_{i}}$ to serve as an expansion mask  \( M^{\text{e}}_i \), which indicates the unseen tissue regions with excessively high opacity (defined by a threshold $\delta=0.8$). Similar to the scene initialization, in the $i$-th frame, SGR is also employed to predict the corresponding Gaussian attributes for expanding areas masked by \( M^{\text{e}}_i \).  Estimated attributes within these areas are further transformed to the world coordinate system with the estimated pose \(\mathbf{T}_i \) to obtain the expanding Gaussian points $\mathcal{G}_{\text{e}}^i$ for unseen regions,
\begin{equation}
    \mathcal{G}_{\text{e}}^i:=tf(M^{\text{e}}_i \odot SGR(\mathbf{I}_i,\mathbf{D}_i),\mathbf{T}_i)
\end{equation}
where $tf(\cdot)$ denotes the camera-to-world transformation, and  $M^{\text{e}}_i$ is rendered by deformed Gaussian $\tilde{\mathcal{G}}^{i-1}_{t_{i}}$. Subsequently, canonical Gaussian model for the $i$-th frame is produced by merging previous canonical model $\mathcal{G}^{i-1}$ and expanded Gaussians $\mathcal{G}_{\text{e}}^i$:
\begin{equation}
    \mathcal{G}^i= \mathcal{G}^{i-1}\oplus \mathcal{G}_{\text{e}}^i
\end{equation}
where $\oplus$ represents the concatenation of Gaussian point clouds. Therefore, the deformed scene at time $t_i$ can be described with the newest Gaussians $ \mathcal{G}^i$ and deformation model, written as $ \tilde{\mathcal{G}}^i_{t_i}= \mathcal{G}^i+\Phi(t_i)$.

\vspace{-3mm}
\subsection{Retrospective Deformation Recapitulation}

\label{sec:3.4}

Despite that existing deformation modeling techniques~\cite{yang2024deform3dgs,cao2023hexplane,liu2024endogaussian,wang2022neural} effectively approach deformable scene reconstruction, simply integrating them into frame-by-frame training framework fails to reach perfect reconstruction quality since newly learned deformations may overwrite previously modeled ones. 
Driven by these issues, we introduce a time-decoupling deformation model (TDDM) with partially learnable basis functions to minimize computational cost and mitigate the risk of newly learned deformations overwriting the prior ones. Furthermore, a new retrospective recapitulation learning strategy is proposed to refine deformations of earlier frames which maintains historical information.

\textbf{Temporal decoupling deformation model.}

Our TDDM $\Phi$ models tissue deformations by utilizing four deformable attributes $\{\mathbf{c}, \mathbf{s}, \mathbf{r}, \bm{\mu}\}$. We define a separate deformation function $\Phi^{attr}(t)$ parameterized by timestamp $t$ for each of them, where $attr\in\{\mathbf{c}, \mathbf{s}, \mathbf{r}, \bm{\mu}\}$. Each deformation model $\Phi$ is defined by a linear combination of $B$ learnable Gaussian basis functions $\{\varphi_j\}$ parameterized by $\Theta=\{\Theta_j\}$, where $\Theta_j$ contains the mean $\tau_j$, variance $\sigma_j^2$, and weight $w_j$ for Gaussian function, illustrated as following:
%Our TDDM $\Phi$ models tissue deformations with multiple learnable Gaussian basis functions $\{\varphi_j\}$ (Fig.~\ref{fig:deformation}). Specifically, we regard $\{\mathbf{c}, \mathbf{s}, \mathbf{r}, \bm{\mu}\}$ as deformable attributes and define a separate deformation function $\Phi^{attr}(t)$ for each of them, where $attr\in\{\mathbf{c}, \mathbf{s}, \mathbf{r}, \bm{\mu}\}$. Our deformation model $\Phi$ is defined by a linear combination of $B$ Gaussian basis functions $\{\varphi_j\}$ defined by learnable parameters $\Theta=\{\Theta_j\}$, where $\Theta_j$ contains the mean $\tau_j$, variance $\sigma_j^2$, and weight $w_j$ for Gaussian function, illustrated as following:
\vspace{-2mm}
\begin{equation}
\Phi(t;\Theta)=\sum_{j=1}^B{ \varphi_j(t;\Theta_j)}
\end{equation}
\vspace{-2mm}
\begin{equation}
\varphi_j(t;\Theta_j)=w_j\exp(-\frac{1}{2\sigma_j^2}(t-\tau_j)^2)
\label{phi}
\end{equation}
where the means $\{\tau_j\}$ are evenly initialized across the entire time span, \( \tau^{\text{init}}_j = t_{\text{max}} \cdot j/(B - 1) \). With Gaussian basis functions, deformations at current time step \yj{$t_i$} are predominantly modeled by $\varphi_j$ with $\tau_j$ close to $t_i$, which mitigates the interference from temporally distant basis functions, effectively decoupling them from current-frame deformation modeling.

\begin{figure}[t]
	\centering
	\includegraphics[width=0.43\textwidth]{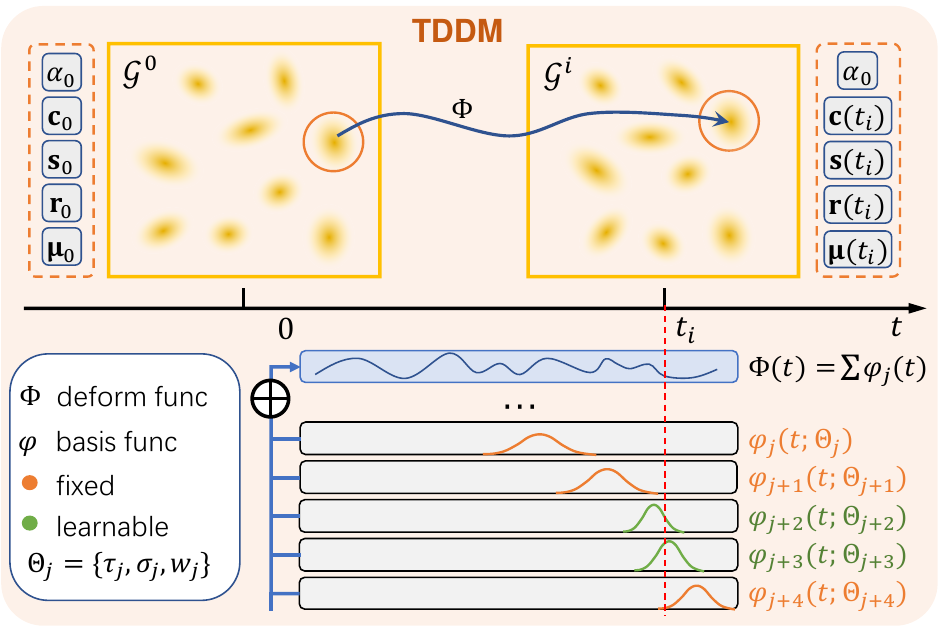}
	\vspace{-3mm}
	\caption{Illustration of the proposed TDDM. Deformation functions \( \Phi(t) \) are defined to represent the attributes' deviation from the canonical values over time. Each one is articulated as an accumulation of Gaussian basis functions \( \{\varphi_j\} \). Only partial basis functions are activated and their parameters are optimized during training.} 
	\label{fig:deformation}
\end{figure}

Meanwhile, we observe that the per-Gaussian deformation modeling approach can become increasingly computationally expensive due to increasing Gaussian point numbers when the scene expands progressively. Given the property that Gaussian functions with mean values $\tau$ distant from current time step $t_i$ exhibit minor influence on current-frame deformations $\Phi(t_i)$, only optimizing basis functions near the current time step can remain comparative reconstruction quality while minimizing computational expense. Therefore, as shown in Fig.~\ref{fig:deformation}, we propose a novel optimization strategy where only a subset of basis functions are optimized for each training iteration while others remain fixed, reducing learnable parameters and thus enhancing training efficiency. Specifically, when modeling the deformation at time \( t_i \), only neighboring $2m$ ($m$ is set to 4 in this work) basis functions are optimized, selected from the subset $\{\varphi_j(t_i; \tau^{init}_j)\;|\; \tau^{init}_j\in(t_i-t_{max}m/(B-1), t_i+t_{max}m/(B-1))\}$.

\textbf{Retrospective recapitulation learning.}

According to our observation, despite utilizing a TDDM designed to \yj{alleviate inter-basis interference}, scene deformations in earlier frames still degrade as training progresses in a frame-by-frame training manner. Since deformations at neighboring time steps may be predominantly contributed by the same group of basis functions, newly learned deformations of these basis functions can still interfere with the previously modeled ones. Herein, after exploiting scene expansion scheme, we propose the retrospective recapitulation learning strategy to further enhance historically learned deformations, maintaining the reconstruction performance of earlier frames. 

As shown in Fig.~\ref{fig:method} (d), retrospective recapitulation learning is performed in per-frame training. Specifically, in the \(i\)-th frame training, the deformation model is retroactively trained on a set of previous frames randomly sampled from the past \(w\) time steps. For a \(h\)-th historical frame in the training set, the tissue deformation \(\Phi(t_h)\) at time \(t_h\) is computed and superimposed onto the latest canonical Gaussian \(\mathcal{G}^i\) to describe the dynamic scene $\tilde{\mathcal{G}}^i_{t_h}=\mathcal{G}^i+\Phi(t_h)$. The previously learned camera pose \(\mathbf{T}_h^K\) for the \(h\)-th frame is utilized for rendering and computing the photometric loss. During this learning phase, only the deformation model is updated. Note that deformations for Gaussian points \(\{\mathcal{G}^k_{\text{e}}| k \in (h, i]\}\), expanded after the $h$-th frame training, may not be optimized, as they are theoretically unobservable under \(\mathbf{T}_h^K\).  Through the proposed retrospective learning, our framework effectively maintains prior scene deformation and thus enhances the reconstruction quality for long-duration surgical video.

\subsection{Objective Functions and Training Setting}
We employ different loss objective functions for the pre-training of SGR network, and the reconstruction phase including joint learning and retrospective recapitulation learning. These functions are designed to minimize the difference between rendered images and depth maps with their ground truth (GT) counterparts, leading to reconstructions with high visual fidelity. 
Besides, benefiting from the differentiable rasterizer~\cite{kerbl20233d,yugay2023gaussian} used for rendering, gradients of loss functions can be efficiently calculated and back-propagated to the learnable parameters including SGR network weights $\Lambda$, TDDM parameters $\Theta$, and camera poses $\mathbf{T}$. 

Specifically, we utilize \( L_1 \) loss for the rendered RGB images and depth maps. For different training stages, specific masks \( M \) are used to indicate the learning regions, with pixels outside these regions excluded from the loss computation:  
\begin{equation}
   L(M)=\left | M\odot(\hat{\mathbf{I}}-\mathbf{I}) \right | + \lambda^\text{D}\left |M\odot (\hat{\mathbf{D}}-\mathbf{D}) \right |
\end{equation}
where \( \lambda^{D} \) is a balancing hyperparameter that scales the contribution of the depth loss.

During SGR pre-training, we employ a full-one mask \( M^{1} \) to predict suitable Gaussian attributes for each pixel, leading to the loss function:  $\mathcal{L}_\text{SRG}=L(M^{1};\Lambda)$. 

During the joint reconstruction and camera pose learning,
for each frame \( i\)-th, images are rendered from the previously reconstructed \( \mathcal{G}^{i-1} \) and may contain unseen regions, denoted by \( M^{\text{e}} \), which are excluded from the loss computation. Additionally, following prior works~\cite{liu2024endogaussian,wang2022neural,yang2024deform3dgs}, a surgical instrument mask $M^{\text{i}}$ is applied to mitigate occlusion effects from instrument. Meanwhile, to prevent deformation model from learning incorrect Gaussian deformations caused by camera movement, an extra regularization term \( \Psi \) is introduced. This term penalizes the average displacement of all Gaussian points, ensuring realistic deformation learning. The overall objective for the joint learning includes both:  
\begin{equation}
    \mathcal{L}_{\text{JL}} = L((1-M^{\text{e}})\odot M^{\text{i}};\Theta,\mathbf{T}) + \Psi(t_i;\Theta)
\end{equation}
\begin{equation}
    \Psi(t_i;\Theta)=\left \|\frac{1}{N}\sum_{n=1}^{N}{\Phi^{\bm{\mu}}_n}(t_i)\right \| ^2
\end{equation}

For the retrospective recapitulation learning,
there is no unseen region but we still aim to prevent the occlusion effect caused by the instrument, for accurate reconstruction. 
Therefore, we set loss function as $\mathcal{L}_\text{RRL}=L(M^\text{i};\Theta)$.

To enhance the applicability of the framework for the reconstruction of long video sequences, we utilize a multi-model representation method, drawing inspiration from \cite{stilz2024flex}. During the iterative frame-by-frame training process, once the number of frames reconstructed by current model exceeds a preset threshold \( \kappa=100 \), the model is re-initialized on the next frame and trained following the same procedure in Fig.~\ref{fig:method}. This method prevents accumulating an excessive number of Gaussians, which could otherwise lead to substantial memory consumption and a consequent decrease in efficiency. 

\section{Experiments}
\subsection{Datasets and Implementation}
\noindent \textbf{StereoMIS dataset.}
We utilize a public benchmark dataset StereoMIS to evaluate our proposed method~\cite{hayoz2023learning}. It is comprised of multiple stereo vivo videos in the da Vinci robotic surgeries on three porcine subjects. Following the selection criteria outlined in \cite{stilz2024flex}, we choose five sequences for reconstruction, each with 1000 frames. The content effectively captures a variety of scenarios that are frequently encountered in surgical procedures. Raw images are of high resolutions of 1024$\times$1280 and we downsample images to a size of 512$\times$640 to further enhance learning efficiency.

\noindent \textbf{Hamlyn datasets.}
We further validate the efficacy of our method on another widely-used public benchmark Hamlyn dataset~\cite{stoyanov2005soft,mountney2010three}. It encompasses a variety of da Vinci robotic surgical sequences from multiple procedures. For this study, we select three sequences, each consisting of 1000 frames, derived from the preprocessed data provided by \cite{recasens2021endo}. These sequences contain a range of complex scenarios, including rapid camera motion, extensive tissue movement, respiratory motion in conjunction with camera movement, and tissue interactions with camera movement. We do a preprocessing of cropping the images to 400$\times$288 to avoid invalid regions in the rectified images.

We employ four widely-used evaluation metrics to quantify the fidelity of the renderings against the GT, i.e., PSNR, SSIM, LPIPS with backbones of AlexNet (LPIPSa) and VGG (LPIPSv). Also, we record the training time and the rendering speed to measure the method efficiency.

\noindent \textbf{Implementation Details.}
Given the significant disparities in video content, surgical operations, and image size between the StereoMIS and Hamlyn datasets, we pre-train two SGR models respectively. For pre-training, 2000 frames of each dataset are randomly selected from various sequences, excluding those used for Free-DyGS training. Each model is trained for 100 epochs and they achieve PSNR of 34.17 and 32.63 on the validation subsets of the StereoMIS and Hamlyn datasets, respectively. 
Once pre-trained, these models were frozen during Free-DyGS training.
In our approach, for each per-frame training process, the joint learning lasts for 10 iterations and the retrospective learning lasts for 40 iterations. Adam optimizer is utilized with an initial deformation learning rate of $1.6e-4$, camera rotation learning rate of $1e-3$, and camera translation learning rate of $0.05$. 
All experimental procedures are conducted on an NVIDIA RTX A5000 GPU.

\vspace{-2mm}
\subsection{Comparison with State-of-the-art Methods}

\begin{figure*}[t]
	\centering
	\includegraphics[width=0.93\textwidth]{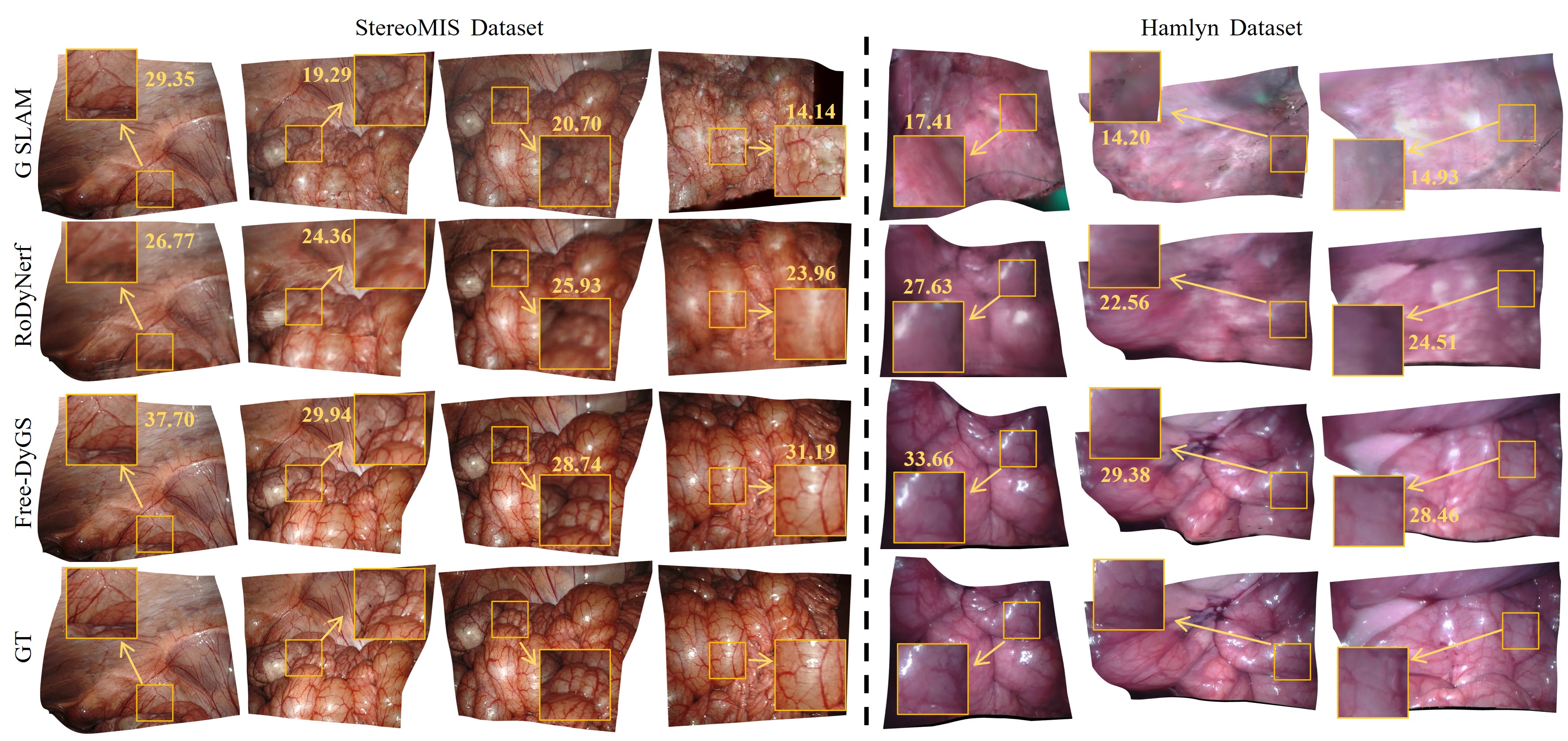}
	\vspace{-3mm}
	\caption{Qualitative comparisons of different methods on typical frames from both datasets. We show the rendering PSNR of corresponding frames.} 
	\label{fig:deform_results}
	\vspace{-5mm}
\end{figure*}

\subsubsection{Results on StereoMIS dataset}
We compare our Free-DyGS against several SOTA methods for camera-pose-free scene reconstruction under moving camera, including Flex~\cite{stilz2024flex}, RoDyNeRF~\cite{liu2023robust} and GSLAM~\cite{yugay2023gaussian}. Flex and RoDyNeRF are both NeRF-based methods specifically designed for dynamic scene reconstruction. Flex addresses general tissue deformations in surgical scenarios, while RoDyNeRF targets moving objects in natural environments. In contrast, GSLAM, based on 3DGS, is proposed to reconstruct static natural scenes. %lacking additional mechanisms to handle scene deformations.
Additionally, we further compare our method with several advanced surgical scene reconstruction approaches, like \cite{stilz2024flex}. Note that most of these methods focus on tissue deformation modeling while assuming a fixed camera setup, including Forplane\cite{yang2023efficient}, LocalRF~\cite{meuleman2023progressively}, and EndoSurf~\cite{zha2023endosurf}. For these methods, we introduced a pre-trained pose estimator RPE~\cite{hayoz2023learning} to allow them for the unposed setting, namely hybrid approaches. 
We follow the experimental setting of Flex \cite{stilz2024flex}, therefore, the results of Flex and other hybrid methods are quoted from the original paper \cite{stilz2024flex}. For GSLAM and RoDyNeRF, we utilize their released code to do the re-implementation. During preprocessing for RoDyNeRF, we utilize the stereo depth estimation instead of the original monocular estimation for fair comparison. 

\begin{table}[t]
\caption{Quantitative results on StereoMIS datasets.}
\vspace{-2mm}
\label{tab:steromis_results}
\centering
\resizebox{0.48\textwidth}{!}{
\begin{threeparttable}
\begin{tabular}{c|ccccc}
\hline
  Method &  PSNR$\uparrow$& SSIM$\uparrow$ & LPIPSv$ \downarrow$& LPIPSa$ \downarrow$& Time$\downarrow$ \\ \hline
   GSLAM~\cite{yugay2023gaussian}& 17.84& 0.520& 0.453& 0.424
&25.7min \\
  RoDyNeRF~\cite{liu2023robust}
& 24.13& 0.629& 0.438& 0.429
&  $>$24H \\
Flex$^*$~\cite{stilz2024flex}& 30.62& 0.818& 0.245& \textbf{0.179}
&20H \\ \hdashline
 RPE+EndoSurf$^*$\cite{zha2023endosurf}& 25.18& 0.622& 0.529& 0.528 &- \\
RPE+LocalRF$^*$\cite{meuleman2023progressively}& 27.41& 0.781& 0.288& 0.245&- \\
 RPE+ForPlane$^*$\cite{yang2023efficient}& 30.35& 0.783& 0.301& 0.208&- \\  \hline
w/o SGR& 31.22& 0.859& 0.232& 0.206 &19.9min \\
 w/o J\&E& 28.95 & 0.792 & 0.276 & 0.257 &\textbf{11.3min} \\
 w/o TDDM& 20.69 & 0.644& 0.419&0.401 & 12.0 min\\
  w HexDM& 27.17& 0.751& 0.334&0.309& 40.8 min\\
 w/o RRL& 26.96& 0.768& 0.308& 0.284 &12.0min \\ \hline
  % Free-DyGS-H& \textbf{31.99} & \textbf{0.871} & \textbf{0.211} & \uline{0.187} &16.2min \\ 
 Free-DyGS(Ours)
& \textbf{31.90}& \textbf{0.870}& \textbf{0.211}&0.187
&11.9min \\ \hline

\end{tabular}
\end{threeparttable}
}
 $^*$ Note: Results are derived from \cite{stilz2024flex}. ~~~~~~~~~~~~~~~~~~~~~~~~~~~~~~~~~
\vspace{-5mm}
\end{table}

Reconstruction results and training times of various methods are presented in Table \ref{tab:steromis_results}. We can see that our approach demonstrates a significant improvement over the static scene reconstruction technique, GSLAM, across all evaluation metrics. Meanwhile, we require only half the training time of GSLAM, likely due to GSLAM's inherent limitations in handling tissue deformation. Compared to RoDyNeRF, our method excels in addressing surgical scenes characterized by complex tissue deformation. Furthermore, our approach surpasses Flex in most evaluation metrics, which also offers a substantial reduction in training time, making a high usability for clinical applications. Additionally, Free-DyGS outperforms hybrid methods that integrate RPE, which have yet to achieve the desired reconstruction accuracy. This discrepancy may be attributed to the fact that such methods are designed for fixed-camera setups and lack robust strategies for scene expansion. As a result, our Free-DyGS achieves superior performance in scene reconstruction on the StereoMIS dataset, peaking a new state-of-the-art with a PSNR of 31.90.
% and the PSNR of 31.99 for Free-DyGS-H, respectively.

\subsubsection{Results on Hamlyn dataset}
We further validate our method on the Hamlyn dataset by first comparing it with advanced unposed reconstruction methods, including RoDyNeRF and GSLAM. Note that Flex~\cite{stilz2024flex} neither reported results on the Hamlyn dataset nor released the code, making it challenging to reimplement and compare with it fairly. Similarly, we also compare the hybrid methods. For surgical scene reconstruction, we compare with Forplane\cite{yang2023efficient} given its promising performance. We also include two advanced GS-based approaches, EndoG~\cite{liu2024endogaussian} and Deform3DGS~\cite{yang2024deform3dgs}. For camera pose estimation, we employ EDaM~\cite{recasens2021endo}, which is specifically designed for Hamlyn dataset. We employ the released codes of these hybrid methods and try the best to tune the hyperparameters.

\begin{table}[t]
\caption{Quantitative results on Hamlyn datasets.}
\vspace{-2mm}
\label{tab:hamlyn_results}
\centering
\resizebox{0.48\textwidth}{!}{
\begin{threeparttable}
\begin{tabular}{c|ccccc}
\hline
  Method &  PSNR$\uparrow$& SSIM$\uparrow$ & LPIPSv $\downarrow$& LPIPSa $\downarrow$& Time$\downarrow$\\ \hline
 GSLAM~\cite{yugay2023gaussian}& 20.60& 0.717& 0.441& 0.387&22.6min\\
  RoDyNeRF~\cite{liu2023robust}
& 26.75& 0.796& 0.354& 0.313&  $>$24H\\ \hdashline
 EDaM+EndoG~\cite{liu2024endogaussian}& 24.36& 0.767& 0.471& 0.423&12.5min\\
 EDaM+Deform3DGS~\cite{yang2024deform3dgs}
& 26.31& 0.829& 0.398& 0.352&9.2min\\
  EDaM+Forplane~\cite{yang2023efficient}
& 26.92& 0.807& 0.419& 0.380&8.6min\\  \hline
 w/o SGR & 29.30& 0.874& 0.289& 0.253 &7.9min\\
 w/o J\&E & 29.63 & 0.879& 0.281& 0.244&\textbf{6.1min}\\
  w/o TDDM& 22.69 & 0.778& 0.459&0.414 & 6.9 min\\
 w HexDM& 27.07& 0.843& 0.374& 0.331&28.8min\\
 w/o RRL& 24.43& 0.766& 0.387& 0.336&6.8min\\ \hline
 % Free-DyGS-H& \textbf{30.03} & \textbf{0.885}& \textbf{0.271} & \uline{0.234} &8.5min\\
 Free-DyGS(Ours)
& \textbf{30.01}& \textbf{0.882}& \textbf{0.271}& \textbf{0.231}&6.6min\\ \hline

\end{tabular}
\end{threeparttable}
}
\vspace{-3mm}
\end{table}

The quantitative results are summarized in Table~\ref{tab:hamlyn_results}, where our approach demonstrates superior performance compared to both GSLAM and RoDyNeRF, due to their limited ability to model complex tissue deformations. Using the camera trajectory estimated by EDaM, the ForPlane, Deform3DGS, and EndoG methods exhibit varying degrees of accuracy, attributed to their differing abilities to model deformations. Our method surpasses all the compared state-of-the-art techniques by a significant margin, in terms of both accuracy and efficiency. Our Free-DyGS achieves a PSNR of 30.01 with training time of 6.6 minutes.

\subsubsection{Qualitative Comparison} 
We conduct a visual comparison on both datasets, with reconstruction results on representative frames presented in Fig.~\ref{fig:deform_results}. It can be observed that GSLAM and RoDyNeRF struggle in challenging cases, often producing blurry and low-quality images. For instances, GSLAM significantly misestimates camera poses, leading to substantial discrepancies between the rendered scenes and the GT. Additionally, dynamic tissue rendering suffers from severe blurring, further degrading the overall reconstruction quality.
While RoDyNeRF can represent a plausible scene structure, it fails to recover fine details accurately. As shown in the zoomed-in view, its rendered results appear blurry, particularly in delicate structures such as vascular tissues. 
In contrast, our Free-DyGS accurately renders the scene and preserves fine details. Our method consistently achieves precise reconstructions with higher PSNR, demonstrating superior performance in both overall quality and detail preservation.

\subsubsection{Time Efficiency} 

We further compare the training time and rendering speed with different methods.  In line with the definitions used in traditional SLAM methods, the overall training time for scene reconstruction under the camera-pose-free setting can be divided into two parts: tracking time and reconstruction time. The former evaluates the time required for camera pose estimation, while the latter assesses the subsequent refinement of the reconstruction. The overall training time for a 1000-frame dynamic surgical sequence and rendering speed are shown in Table~\ref{tab:training_time}.

Regarding training time, RoDyNeRF, a NeRF-based approach, exhibits a significantly longer reconstruction time than other methods, exceeding 24 hours. 
GSLAM employs a multithreading paradigm to accelerate the process, yet its total reconstruction time still amounts to 22 minutes due to its lack of deformation modeling capability. 
Our experiments reveal that hybrid methods demonstrate shorter reconstruction times and Deform3DGS and EndoG require longer reconstruction times compared to ForPlane, since GS-based methods rely on cloning and splitting to densify Gaussian points with more training iterations under camera motions. In contrast, ForPlane benefits from the implicit representation of NeRF, obviating the need for this explicit densification.
Our Free-DyGS simultaneously performs camera tracking and dynamic scene reconstruction in a frame-by-frame manner. It achieves the most efficient reconstruction with 6.6 minutes per surgical sequence.
Regarding rendering speed, our method, along with other GS-based approaches, reach real-time performance exceeding 100 FPS. In contrast, RoDyNeRF and ForPlane exhibit significantly lower rendering speeds with only 0.67 FPS and 1.56 FPS, respectively. 
Notably, only our Free-DyGS and GSLAM are trained in a frame-by-frame manner which cater for online surgical applications, whereas others necessitate entire videos for training.

\begin{table}[t]
\caption{Training time and rendering speed on the Hamlyn dataset.}
\vspace{-2mm}
\label{tab:training_time}
\centering
\resizebox{0.48\textwidth}{!}{
\begin{threeparttable}
\begin{tabular}{c|cccc}
\hline
  Method & \makecell[c]{Tracking\\time$\downarrow$}  & \makecell[c]{Reconstruction\\time$\downarrow$}& \makecell[c]{Total\\time$\downarrow$}& \makecell[c]{Rendering\\speed (FPS)$\uparrow$}\\ \hline
  RoDyNeRF~\cite{liu2023robust}
& $>$10H& $>$10H& $>$24H& 0.67\\
 GSLAM~\cite{yugay2023gaussian}& 20.5min& 20.9min& 22.6min& \textbf{100+}\\ \hdashline
 % EDaM\cite{recasens2021endo}& 4.8min& -& -&-\\
 EDaM+Deform3DGS~\cite{yang2024deform3dgs}& 4.8min& 4.5min& 9.2min& \textbf{100+}\\
 EDaM+EndoG\cite{liu2024endogaussian}& 4.8min& 7.7min& 12.5min& \textbf{100+}\\
 EDaM+Forplane\cite{yang2023efficient}& 4.8min& \textbf{3.8min}& 8.6min& 1.56\\ \hline
 Free-DyGS(Ours)
& \textbf{1.5min}&5.1min& \textbf{6.6min}& \textbf{100+}\\ \hline

\end{tabular}
\end{threeparttable}
}
\vspace{-5mm}
\end{table}

\vspace{-2mm}
\subsection{Ablation Study on Key Components}
\label{sec:4.3}
%这里的名字，是不是要按方法部分的update一下
%Joint Learning这块这能不能再formulate一下，keyword可以用Synergistic，最好也把expansion融合进来
To validate the efficacy of the key components proposed in our method, we conduct ablation experiments under five configurations: (i) Without SGR Initialization (w/o SGR): We initially parameterize Gaussians with default values in GS-based methods. (ii) Without Joint Learning and Scene Expansion (w/o J\&E): We sequentially optimize the camera pose and deformation model in each iteration. 
(iii) Without TDDM (w/o TDDM): We omit the deformation model and optimize the Gaussian attributes in canonical space during training. (iv) With Hexplane as the Deformation Model (w/ HexDM): We replace the TDDM in our framework with the widely adopted Hexplane deformation model. (v) Without Retrospective Recapitulation Learning (w/o RRL): We omit the retrospective learning process and focus solely on learning the deformation of the current frame. 

Table~\ref{tab:steromis_results} and Table~\ref{tab:hamlyn_results} present the comparative results of our full model Free-DyGS against other ablation settings on StereoMIS and Hamlyn datasets, respectively. 
We observe that without SGR which efficiently parameterizes initial Gaussians, the model requires additional training iterations to optimize the initial Gaussian parameters, leading to much longer training time with inferior rendering quality. When comparing the results of w/o J\&E, Free-DyGS demonstrates superior rendering quality across both datasets, indicating our synergistic training strategy can effectively capture a strong correlation between camera pose estimation and scene deformation modeling. As expected, the results of w/o TDDM are poor on both datasets, given that the lack of dynamic scene modeling creates a vicious cycle of poor reconstruction and inaccurate pose estimation. Different from TDDM, Hexplane employs a spatial-temporal structure encoder and incorporates total variation loss to smooth the spatiotemporal deformation model, resulting in significant temporal coupling. Although Hexplane has demonstrated strong performance in numerous existing studies \cite{liu2024endogaussian, yang2023neural, wu20244d}, it exhibits more pronounced limitations during frame-by-frame training compared to our TDDM.
Furthermore, we see that results of w/o RRL are substantially lower than our full model given the same training iteration numbers. This setting focuses solely on learning the current-frame deformations without guidance from historical frames, largely degrading previously modeled deformations.

\vspace{-3mm}
\subsection{Detailed Analysis of RDR learning}
\label{sec:4.5}

Retrospective deformation recapitulation learning is a key component of our framework, which balances newly acquired deformation with knowledge learned from previous frames. During this phase, the deformation model is retroactively trained on frames randomly sampled from a preceding window of $\omega$ frames. In this section, we delve into the impact of this critical window width $\omega$ by conducting experiments with varying window widths $\omega\in\{40,70,150,200\}$. Also, we validate the performance of RDR when sampling from all previous frames to the first ($\omega=\;$toF). The results are summarized in Table~\ref{tab:sampling_results}. 

We observe that as the sampling window width expands from 40 to 100, there is a noticeable enhancement in rendering quality with PSNR increasing from 31.10 to 31.90.
The underlying reason is that increasing the sampling window width properly encourages the model to retain more historical information, thereby alleviating the temporal overwriting of learned information. %应该讨论完 - 
Nevertheless, when the window width reaches 200, the performances slightly decrease and when it further increases to all previous frames, the results drop to a lower PSNR of 30.57, since distant historical moments relatively independent from the current frame are less contributive. Hence, $\omega$ is set as 100 in our framework.

\begin{table}[h]
\caption{Results on StereoMIS with different sampling window widths during retrospective learning}
\vspace{-2mm}
\label{tab:sampling_results}
\centering
\resizebox{0.48\textwidth}{!}{
\begin{threeparttable}
\begin{tabular}{c|cccc}
\hline
  Width &  PSNR$\uparrow$& SSIM$\uparrow$ & LPIPSv$\downarrow$ & LPIPSa $\downarrow$\\ \hline
  $\omega$=40& 31.10 & 0.861 & 0.220 & 0.196 
\\
 $\omega$=70& 31.62 & 0.867 & 0.215 & 0.191 
\\
 $\omega$=100& \textbf{31.90}& 0.870& \textbf{0.211}& \textbf{0.187}
\\
 $\omega$=150& 31.89 & \textbf{0.871} & 0.212 & \textbf{0.187}
\\
 $\omega$=200& 31.85& 0.870& 0.212 & 0.189\\ 
 $\omega$=toF & 30.57& 0.831& 0.245& 0.222
\\\hline

\end{tabular}
\end{threeparttable}
}~~~~~~~~~~~~~~~~~~~~~~~~~~~~
\vspace{-7mm}
\end{table}

\section{Discussion}
Reconstructing scenes from surgical videos is fundamentally crucial for supporting several downstream tasks, such as remote-assisted surgery, surgical navigation and education. However, existing methods struggle to reconstruct scenes involving both dynamic camera poses and deformable tissues, which largely limits their usability in real-world surgical practice where such scenarios commonly appear. %face several limitations, including low quality, the necessity for precise camera poses, and accurate tissue deformation modeling. 
To approach this challenging task, we develop a novel framework, Free-DyGS, introducing GS technology to camera-pose-free reconstruction of dynamic surgical scenes. Our method is trained in a frame-by-frame manner and simultaneously learns camera pose and scene deformation, leading to significant improvements in both reconstruction quality and speed compared to SOTA methods.

In addition, we also conducted some interesting experiments on the StereoMIS dataset. 
Inspired by SLAM methods, we design the randomly sampling retrospective learning phase to refine the reconstruction, which is particularly designed for balancing deformation modeling performances across all frames. In contrast, static scene SLAM methods prioritize utilizing historical information to improve the current reconstruction. This distinction results in different frames sampling strategies for refinement. Our approach employs evenly distributed random sampling, while static scene methods assign varying sampling probabilities to frames. We explore the sampling approach used in EndoGSLAM \cite{wang2024endogslam}, a method designed for static scene reconstruction, where sampling probabilities are determined by the temporal and spatial distances between historical and current frames. However, this approach results in a suboptimal PSNR of 29.43. This indicates that a balanced retrospective strategy is better suited for dynamic scene reconstruction.

To further explore the camera pose estimation task, we also delineate the Absolute Trajectory Errors (ATE), which quantifies the discrepancy between the estimated camera trajectory and the GT. Free-DyGS achieves an ATE of 3.15 mm which is slightly worse than that of Flex (2.57 mm), but markedly superior to other SOTA methods, such as RoDyNeRF (10.22 mm) and GSLAM (23.45 mm).

Although we contribute a novel unposed reconstruction method for deformable surgical scene, there are still some rooms left for further exploration.
Firstly, assuming a stereo surgical scene where the depth map can be derived from stereo depth estimation, our framework cannot be directly applied to monocular surgical videos.
In future research, we will integrate self-supervised monocular depth estimation techniques to empower Free-DyGS to effectively handle monocular videos. 
Besides, despite significantly reduced training time, a 1000-frame surgical video still takes several minutes for reconstruction. We will consider selecting keyframes with significant camera movement or tissue deformation and representing the scene using sparse Gaussians to further increase the efficiency and promote its practical application in intraoperative surgical navigation and remote-assisted surgery in the future.

\section{Conclusion}
In this study, we introduce a novel framework, Free-DyGS, for surgical scene reconstruction with unknown camera motions and complex tissue deformations. We propose a joint learning strategy to simultaneously estimate camera poses and tissue deformations through iterative optimization. We incorporate a pre-trained SGR which effectively parameterizes the initial and expanded Gaussians. A retrospective recapitulation learning strategy is then introduced to address deformation overwriting during frame-by-frame training. Extensive experimental evaluation on two representative surgical datasets underscores the superior performance of our Free-DyGS to existing SOTA methods, demonstrating a notable reduction in training time and an enhancement in rendering quality.

\bibliographystyle{IEEEtran}
\small\bibliography{refs}

% Generated by IEEEtran.bst, version: 1.13 (2008/09/30)
\begin{thebibliography}{10}
\providecommand{\url}[1]{#1}
\csname url@samestyle\endcsname
\providecommand{\newblock}{\relax}
\providecommand{\bibinfo}[2]{#2}
\providecommand{\BIBentrySTDinterwordspacing}{\spaceskip=0pt\relax}
\providecommand{\BIBentryALTinterwordstretchfactor}{4}
\providecommand{\BIBentryALTinterwordspacing}{\spaceskip=\fontdimen2\font plus
\BIBentryALTinterwordstretchfactor\fontdimen3\font minus \fontdimen4\font\relax}
\providecommand{\BIBforeignlanguage}[2]{{%
\expandafter\ifx\csname l@#1\endcsname\relax
\typeout{** WARNING: IEEEtran.bst: No hyphenation pattern has been}%
\typeout{** loaded for the language `#1'. Using the pattern for}%
\typeout{** the default language instead.}%
\else
\language=\csname l@#1\endcsname
\fi
#2}}
\providecommand{\BIBdecl}{\relax}
\BIBdecl

\bibitem{bianchi2020impact}
L.~Bianchi, U.~Barbaresi, L.~Cercenelli, B.~Bortolani, C.~Gaudiano, F.~Chessa, A.~Angiolini, S.~Lodi, A.~Porreca, F.~M. Bianchi \emph{et~al.}, ``The impact of 3d digital reconstruction on the surgical planning of partial nephrectomy: a case-control study. still time for a novel surgical trend?'' \emph{Clinical Genitourinary Cancer}, vol.~18, no.~6, pp. e669--e678, 2020.

\bibitem{schiavina2019three}
R.~Schiavina, L.~Bianchi, M.~Borghesi, F.~Chessa, L.~Cercenelli, E.~Marcelli, and E.~Brunocilla, ``Three-dimensional digital reconstruction of renal model to guide preoperative planning of robot-assisted partial nephrectomy.'' \emph{International Journal of Urology}, vol.~26, no.~9, 2019.

\bibitem{jin2022exploring}
Y.~Jin, Y.~Yu, C.~Chen, Z.~Zhao, P.-A. Heng, and D.~Stoyanov, ``Exploring intra-and inter-video relation for surgical semantic scene segmentation,'' \emph{IEEE TMI}, vol.~41, no.~11, pp. 2991--3002, 2022.

\bibitem{otsuki2023surgical}
Y.~Otsuki, T.~Nuri, E.~Yoshida, K.~Fujiwara, and K.~Ueda, ``Surgical assistant-friendly breast reconstruction using a head-mounted wireless camera with an integrated led light as an educational tool,'' \emph{Plastic and Reconstructive Surgery--Global Open}, vol.~11, no.~4, p. e4940, 2023.

\bibitem{lange2000virtual}
T.~Lange, D.~J. Indelicato, and J.~M. Rosen, ``Virtual reality in surgical training,'' \emph{Surgical oncology clinics of North America}, vol.~9, no.~1, pp. 61--79, 2000.

\bibitem{tang2018augmented}
R.~Tang, L.-F. Ma, Z.-X. Rong, M.-D. Li, J.-P. Zeng, X.-D. Wang, H.-E. Liao, and J.-H. Dong, ``Augmented reality technology for preoperative planning and intraoperative navigation during hepatobiliary surgery: A review of current methods,'' \emph{Hepatobiliary \& Pancreatic Diseases International}, vol.~17, no.~2, pp. 101--112, 2018.

\bibitem{chen2018slam}
L.~Chen, W.~Tang, N.~W. John, T.~R. Wan, and J.~J. Zhang, ``Slam-based dense surface reconstruction in monocular minimally invasive surgery and its application to augmented reality,'' \emph{Computer methods and programs in biomedicine}, vol. 158, pp. 135--146, 2018.

\bibitem{zha2023endosurf}
R.~Zha, X.~Cheng, H.~Li, M.~Harandi, and Z.~Ge, ``Endosurf: Neural surface reconstruction of deformable tissues with stereo endoscope videos,'' in \emph{MICCAI}.\hskip 1em plus 0.5em minus 0.4em\relax Springer, 2023, pp. 13--23.

\bibitem{huang2024endo}
Y.~Huang, B.~Cui, L.~Bai, Z.~Guo, M.~Xu, and H.~Ren, ``Endo-4dgs: Distilling depth ranking for endoscopic monocular scene reconstruction with 4d gaussian splatting,'' in \emph{MICCAI}.\hskip 1em plus 0.5em minus 0.4em\relax Springer, 2024.

\bibitem{liu2024endogaussian}
Y.~Liu, C.~Li, C.~Yang, and Y.~Yuan, ``Endogaussian: Gaussian splatting for deformable surgical scene reconstruction,'' \emph{arXiv preprint arXiv:2401.12561}, 2024.

\bibitem{wang2024endogslam}
K.~Wang, C.~Yang, Y.~Wang, S.~Li, Y.~Wang, Q.~Dou, X.~Yang, and W.~Shen, ``Endogslam: Real-time dense reconstruction and tracking in endoscopic surgeries using gaussian splatting,'' in \emph{MICCAI}.\hskip 1em plus 0.5em minus 0.4em\relax Springer, 2024, pp. 219--229.

\bibitem{guo2024free}
J.~Guo, J.~Wang, D.~Kang, W.~Dong, W.~Wang, and Y.-h. Liu, ``Free-surgs: Sfm-free 3d gaussian splatting for surgical scene reconstruction,'' in \emph{MICCAI}.\hskip 1em plus 0.5em minus 0.4em\relax Springer, 2024, pp. 350--360.

\bibitem{yugay2023gaussian}
V.~Yugay, Y.~Li, T.~Gevers, and M.~R. Oswald, ``Gaussian-slam: Photo-realistic dense slam with gaussian splatting,'' \emph{arXiv preprint arXiv:2312.10070}, 2023.

\bibitem{stilz2024flex}
F.~P. Stilz, M.~A. Karaoglu, F.~Tristram, N.~Navab, B.~Busam, and A.~Ladikos, ``Flex: Joint pose and dynamic radiance fields optimization for stereo endoscopic videos,'' \emph{arXiv preprint arXiv:2403.12198}, 2024.

\bibitem{schoenberger2016sfm}
J.~L. Sch\"{o}nberger and J.-M. Frahm, ``Structure-from-motion revisited,'' in \emph{CVPR}, 2016.

\bibitem{taketomi2017visual}
T.~Taketomi, H.~Uchiyama, and S.~Ikeda, ``Visual slam algorithms: A survey from 2010 to 2016,'' \emph{IPSJ transactions on computer vision and applications}, vol.~9, pp. 1--11, 2017.

\bibitem{zhou2019real}
H.~Zhou and J.~Jagadeesan, ``Real-time dense reconstruction of tissue surface from stereo optical video,'' \emph{IEEE transactions on medical imaging}, vol.~39, no.~2, pp. 400--412, 2019.

\bibitem{song2017dynamic}
J.~Song, J.~Wang, L.~Zhao, S.~Huang, and G.~Dissanayake, ``Dynamic reconstruction of deformable soft-tissue with stereo scope in minimal invasive surgery,'' \emph{IEEE Robotics and Automation Letters}, vol.~3, no.~1, pp. 155--162, 2017.

\bibitem{zhou2021emdq}
H.~Zhou and J.~Jayender, ``Emdq-slam: Real-time high-resolution reconstruction of soft tissue surface from stereo laparoscopy videos,'' in \emph{MICCAI}.\hskip 1em plus 0.5em minus 0.4em\relax Springer, 2021, pp. 331--340.

\bibitem{mildenhall2021nerf}
B.~Mildenhall, P.~P. Srinivasan, M.~Tancik, J.~T. Barron, R.~Ramamoorthi, and R.~Ng, ``Nerf: Representing scenes as neural radiance fields for view synthesis,'' \emph{Communications of the ACM}, vol.~65, no.~1, pp. 99--106, 2021.

\bibitem{wang2022neural}
Y.~Wang, Y.~Long, S.~H. Fan, and Q.~Dou, ``Neural rendering for stereo 3d reconstruction of deformable tissues in robotic surgery,'' in \emph{MICCAI}.\hskip 1em plus 0.5em minus 0.4em\relax Springer, 2022, pp. 431--441.

\bibitem{yang2023neural}
C.~Yang, K.~Wang, Y.~Wang, X.~Yang, and W.~Shen, ``Neural lerplane representations for fast 4d reconstruction of deformable tissues,'' in \emph{MICCAI}.\hskip 1em plus 0.5em minus 0.4em\relax Springer, 2023, pp. 46--56.

\bibitem{yang2023efficient}
C.~Yang, K.~Wang, Y.~Wang, Q.~Dou, X.~Yang, and W.~Shen, ``Efficient deformable tissue reconstruction via orthogonal neural plane,'' \emph{IEEE Transactions on Medical Imaging}, 2024.

\bibitem{kerbl20233d}
B.~Kerbl, G.~Kopanas, T.~Leimk{\"u}hler, and G.~Drettakis, ``3d gaussian splatting for real-time radiance field rendering.'' \emph{ACM Trans. Graph.}, vol.~42, no.~4, pp. 139--1, 2023.

\bibitem{yang2024deform3dgs}
S.~Yang, Q.~Li, D.~Shen, B.~Gong, Q.~Dou, and Y.~Jin, ``Deform3dgs: Flexible deformation for fast surgical scene reconstruction with gaussian splatting,'' in \emph{MICCAI}.\hskip 1em plus 0.5em minus 0.4em\relax Springer, 2024, pp. 132--142.

\bibitem{wang2021nerf}
Z.~Wang, S.~Wu, W.~Xie, M.~Chen, and V.~A. Prisacariu, ``Nerf--: Neural radiance fields without known camera parameters,'' \emph{arXiv preprint arXiv:2102.07064}, 2021.

\bibitem{lin2021barf}
C.-H. Lin, W.-C. Ma, A.~Torralba, and S.~Lucey, ``Barf: Bundle-adjusting neural radiance fields,'' in \emph{ICCV}, 2021, pp. 5741--5751.

\bibitem{bian2023nope}
W.~Bian, Z.~Wang, K.~Li, J.-W. Bian, and V.~A. Prisacariu, ``Nope-nerf: Optimising neural radiance field with no pose prior,'' in \emph{CVPR}, 2023, pp. 4160--4169.

\bibitem{Fu_2024_CVPR}
Y.~Fu, S.~Liu, A.~Kulkarni, J.~Kautz, A.~A. Efros, and X.~Wang, ``Colmap-free 3d gaussian splatting,'' in \emph{CVPR}, June 2024, pp. 20\,796--20\,805.

\bibitem{yan2024gs}
C.~Yan, D.~Qu, D.~Xu, B.~Zhao, Z.~Wang, D.~Wang, and X.~Li, ``Gs-slam: Dense visual slam with 3d gaussian splatting,'' in \emph{CVPR}, 2024, pp. 19\,595--19\,604.

\bibitem{li2024sgs}
M.~Li, S.~Liu, H.~Zhou, G.~Zhu, N.~Cheng, T.~Deng, and H.~Wang, ``Sgs-slam: Semantic gaussian splatting for neural dense slam,'' in \emph{ECCV}.\hskip 1em plus 0.5em minus 0.4em\relax Springer, 2025, pp. 163--179.

\bibitem{liu2023robust}
Y.-L. Liu, C.~Gao, A.~Meuleman, H.-Y. Tseng, A.~Saraf, C.~Kim, Y.-Y. Chuang, J.~Kopf, and J.-B. Huang, ``Robust dynamic radiance fields,'' in \emph{CVPR}, 2023, pp. 13--23.

\bibitem{saha2023based}
S.~Saha, S.~Liu, S.~Lin, J.~Lu, and M.~Yip, ``Based: Bundle-adjusting surgical endoscopic dynamic video reconstruction using neural radiance fields,'' \emph{arXiv preprint arXiv:2309.15329}, 2023.

\bibitem{wang2023bad}
P.~Wang, L.~Zhao, R.~Ma, and P.~Liu, ``Bad-nerf: Bundle adjusted deblur neural radiance fields,'' in \emph{CVPR}, 2023, pp. 4170--4179.

\bibitem{cao2023hexplane}
A.~Cao and J.~Johnson, ``Hexplane: A fast representation for dynamic scenes,'' in \emph{CVPR}, 2023, pp. 130--141.

\bibitem{hayoz2023learning}
M.~Hayoz, C.~Hahne, M.~Gallardo, D.~Candinas, T.~Kurmann, M.~Allan, and R.~Sznitman, ``Learning how to robustly estimate camera pose in endoscopic videos,'' \emph{International journal of computer assisted radiology and surgery}, vol.~18, no.~7, pp. 1185--1192, 2023.

\bibitem{stoyanov2005soft}
D.~Stoyanov, G.~P. Mylonas, F.~Deligianni, A.~Darzi, and G.~Z. Yang, ``Soft-tissue motion tracking and structure estimation for robotic assisted mis procedures,'' in \emph{MICCAI}.\hskip 1em plus 0.5em minus 0.4em\relax Springer, 2005, pp. 139--146.

\bibitem{mountney2010three}
P.~Mountney, D.~Stoyanov, and G.-Z. Yang, ``Three-dimensional tissue deformation recovery and tracking,'' \emph{IEEE Signal Processing Magazine}, vol.~27, no.~4, pp. 14--24, 2010.

\bibitem{recasens2021endo}
D.~Recasens, J.~Lamarca, J.~M. F{\'a}cil, J.~Montiel, and J.~Civera, ``Endo-depth-and-motion: Reconstruction and tracking in endoscopic videos using depth networks and photometric constraints,'' \emph{IEEE Robotics and Automation Letters}, vol.~6, no.~4, pp. 7225--7232, 2021.

\bibitem{meuleman2023progressively}
A.~Meuleman, Y.-L. Liu, C.~Gao, J.-B. Huang, C.~Kim, M.~H. Kim, and J.~Kopf, ``Progressively optimized local radiance fields for robust view synthesis,'' in \emph{CVPR}, 2023, pp. 16\,539--16\,548.

\bibitem{wu20244d}
G.~Wu, T.~Yi, J.~Fang, L.~Xie, X.~Zhang, W.~Wei, W.~Liu, Q.~Tian, and X.~Wang, ``4d gaussian splatting for real-time dynamic scene rendering,'' in \emph{CVPR}, 2024, pp. 20\,310--20\,320.

\end{thebibliography}

\end{document}